\documentclass[11pt,a4paper]{article}
\usepackage[hyperref]{acl2020}
\usepackage{times}
\usepackage{latexsym}
\usepackage{graphicx}

\usepackage{microtype}

\aclfinalcopy 

\title{Encodings of Source Syntax: {S}imilarities in {NMT} Representations Across Target Languages}

\author{Tyler A. Chang \\
  Carleton College \\
  Northfield, MN \\
  \texttt{changt@carleton.edu} \\\And
  Anna N. Rafferty \\
  Carleton College \\
  Northfield, MN \\
  \texttt{arafferty@carleton.edu} \\}

\date{}

\begin{document}
\maketitle
\begin{abstract}
We train neural machine translation (NMT) models from English to six target languages, using NMT encoder representations to predict ancestor constituent labels of source language words.
We find that NMT encoders learn similar source syntax regardless of NMT target language, relying on explicit morphosyntactic cues to extract syntactic features from source sentences.
Furthermore, the NMT encoders outperform RNNs trained directly on several of the constituent label prediction tasks, suggesting that NMT encoder representations can be used effectively for natural language tasks involving syntax.
However, both the NMT encoders and the directly-trained RNNs learn substantially different syntactic information from a probabilistic context-free grammar (PCFG) parser.
Despite lower overall accuracy scores, the PCFG often performs well on sentences for which the RNN-based models perform poorly, suggesting that RNN architectures are constrained in the types of syntax they can learn.
\end{abstract}

\section{Introduction}
Neural machine translation (NMT) encoder representations have been used successfully for cross-task and cross-lingual transfer learning in a variety of natural language contexts (\citealp{Eriguchi2018ZeroShotCC}; \citealp{McCann2017LearnedIT}; \citealp{neubig-hu-2018-rapid}).
Previous work has investigated whether these representations encode syntactic information \citep{shi-etal-2016-string}, as syntactic information is useful in many natural language tasks (\citealp{chen-etal-2017-improved}; \citealp{punyakanok-etal-2008-importance}).
The deep recurrent neural network (RNN) architectures used by many NMT encoders can learn syntactic features, even without explicit supervision (\citealp{blevins-etal-2018-deep}; \citealp{futrell-etal-2019-neural}); NMT encoders specifically have been found to encode information about ancestor constituent labels for words \citep{blevins-etal-2018-deep} and even full syntactic parses of source language sentences \citep{shi-etal-2016-string}.

Cross-linguistically, there is mixed evidence for how target language impacts the encoding of information in NMT encoder representations.
\citet{kudugunta2019investigating} found that representations clustered based on target language family when sentence representations were aligned in a shared space.
However, \citet{belinkov-etal-2017-evaluating} found only small effects of target language on the ability of NMT encoder states to predict part-of-speech (POS) tags.
Because POS tags are typically reliant only on local features within sentences, these contrasting results could suggest that (1) localized encoded information is independent of NMT target language, or (2) encoded syntactic information in general is independent of NMT target language.
In this work, we address the second possibility.

To evaluate more global syntactic information in NMT encoder representations, we assess the ability of NMT encoder states to predict ancestor constituent labels of words; this task is adopted from \citet{blevins-etal-2018-deep}.
Extending \citet{blevins-etal-2018-deep}, we train NMT models towards multiple target languages and evaluate performance on individual constituent labels (e.g. noun phrases).
We find that significant syntactic information is encoded regardless of target language, and target language has little impact on the syntactic information learned by NMT encoders.
Furthermore, we find that NMT encoders rely on explicit morphosyntactic cues to extract syntactic information from sentences.

Finally, by training deep RNNs directly on the constituent label prediction task, we find that RNNs with explicit syntactic training data learn similar syntax to the NMT encoders.
In contrast, a probabilistic context-free grammar (PCFG) parser performs significantly differently from both RNN-based models, suggesting that RNNs may be constrained by their reliance on explicit syntactic cues.

\section{Methodology}
We trained NMT models from English to six different target languages, assessing the ability of NMT encoder states to predict POS, parent, grandparent, and great-grandparent constituent labels of words.

\subsection{NMT Models}
NMT models were trained on the United Nations (UN) Parallel Corpus, using the fully aligned subcorpus of approximately 11 million sentences translated to all six UN official languages: English, Spanish, French, Russian, Arabic, and Chinese \citep{ziemski-etal-2016-united}.
NMT models were trained from English to each target language using OpenNMT’s PyTorch implementation \citep{klein-etal-2017-opennmt} with byte pair encoding for subword tokenization in all languages \citep{sennrich-etal-2016-neural}.
Each NMT encoder and decoder was a unidirectional four-layer long short-term memory (LSTM; \citealp{hochreiter1997lstm}) network with 500 dimensions, using dot-product global attention in the decoder \citep{luong-etal-2015-effective}.
Each NMT model was trained for 11 epochs (approximately 2,000,000 steps) using Adam optimization \citep{kingma2014adam}.\footnote{
The first 10 epochs used learning rate 0.0002; the learning rate was halved every 30,000 steps during the final epoch.
}
The model with the best performance on the UN evaluation dataset for each language was used to generate encoder representations in the constituent label prediction task.

\subsection{Constituent Label Predictions}
\label{sec:PredictionModels}
\paragraph{Dataset}
Constituent label predictions used tree-parsed sentences from the CoNLL-2012 dataset, containing sentences from English news and magazine articles, web data, and transcribed conversational speech \citep{pradhan-etal-2012-conll}.

As in \citet{blevins-etal-2018-deep}, constituent label models were trained on the CoNLL-2012 development dataset and tested on the test dataset.  A subset of the CoNLL-2012 training dataset was used as an evaluation dataset; the training, evaluation, and test datasets each contained approximately 160,000 English words.

\paragraph{Prediction models}
We trained simple feedforward neural networks to predict ancestor constituent labels (POS, parent, grandparent, and great-grandparent) of words, given the NMT encoder state after reading the word.
The NMT encoders were kept fixed during constituent label training.
We used the deepest encoder layer as our encoder representation; deeper layers have been shown to perform better on constituent label prediction tasks \citep{blevins-etal-2018-deep}.

Each feedforward network contained one 500-dimensional hidden layer, and each model was trained until it completed 10 consecutive epochs with no improvement on the evaluation dataset.
To account for variation between models based on random initialization of weights and shuffling of the training data, we trained 20 feedforward models for each combination of NMT encoder target language and constituent label (POS, parent, grandparent, or great-grandparent).

\paragraph{Baselines}
We computed a baseline accuracy for each constituent label prediction task by simply predicting the most frequent constituent label given the current input word (e.g. given the current input word ``dog,'' the most frequent POS tag would be NN for ``singular noun'').  This baseline accuracy is the maximum possible accuracy for a deterministic model that only knows the current input word.

\section{Results}
\begin{figure}
    \centering
    \includegraphics[width=7.7cm]{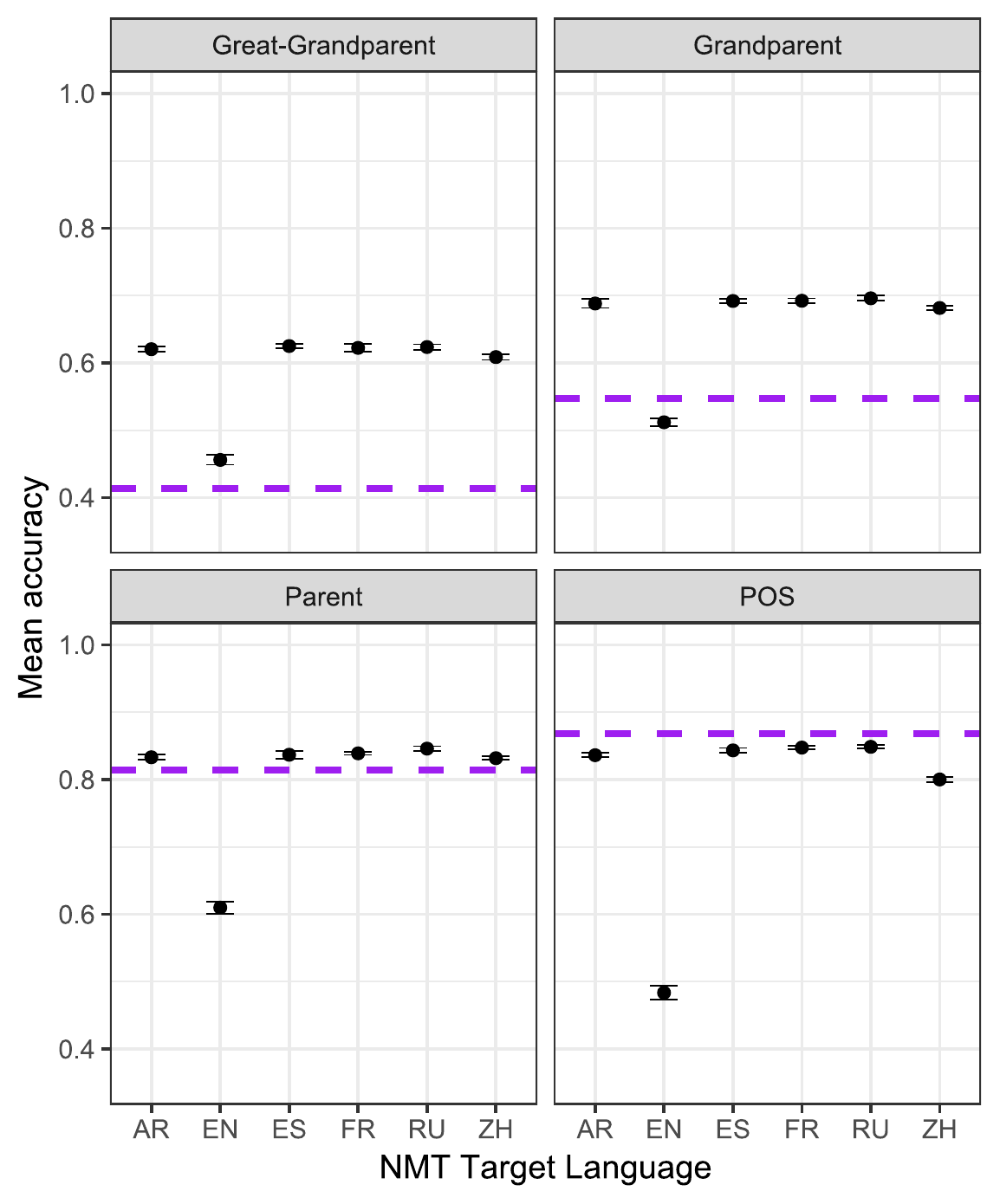}
    \caption{Results for the constituent label prediction tasks, trained from NMT encoder representations.
    Dots indicate mean accuracies (based on 20 feedforward models), bars indicate two standard deviations from the mean, and dashed lines represent baseline accuracies.}
    \label{fig:LanguageResults}
\end{figure}

\paragraph{NMT encoders learned syntax.}
As shown in Figure \ref{fig:LanguageResults}, NMT encoder representations for all target languages except the autoencoder resulted in accuracy scores above the baseline for the parent, grandparent, and great-grandparent constituent label tasks (adjusted $p < 0.001$ for all comparisons, using one sample $t$-tests).
The English autoencoder was the only target language without consistent performance above the baselines for these tasks; NMT autoencoders have been found to memorize sentences without learning syntactic information \citep{shi-etal-2016-string}.
These results indicate that with the exception of autoencoders, NMT encoder representations contain syntactic information regardless of target language.

\paragraph{Models performed poorly for POS tags.}
In contrast to \citet{blevins-etal-2018-deep} but in line with \citet{belinkov-etal-2017-evaluating}, all target languages performed slightly below the baseline for the POS prediction task (adjusted $p < 0.001$ for all comparisons, using one sample $t$-tests).
This result may be because POS encodes less useful information than other features for machine translation tasks.
For instance, \citet{belinkov-etal-2017-evaluating} found that models performed above the baseline if the task was modified to predict semantic tags.

\subsection{Similarities Across Target Languages}
While there were statistically significant differences in accuracy between target languages for all four constituent label tasks (one-way ANOVA, $p < 0.001$ for all tasks), these differences were quite small.
The non-English target languages varied by less than 2\% within each of the parent, grandparent, and great-grandparent constituent label tasks (see Figure \ref{fig:LanguageResults}).

\paragraph{NMT encoders learned similar syntax.}
To further test the hypothesis of similar syntactic information across encoder representations, we assessed the performance of the NMT encoders on individual constituent labels (e.g. noun phrases).
To do this, we considered the constituent label predictions as the results of a binary classification task for each individual label.
For instance, when considering the noun POS tag, all POS tags were separated into two categories: noun and not noun.
Then, we computed F1 scores for individual constituent labels for each NMT model, allowing us to quantify similarities between NMT encoders based on individual label performance.

\begin{figure*}
    \centering
    \includegraphics[width=16cm]{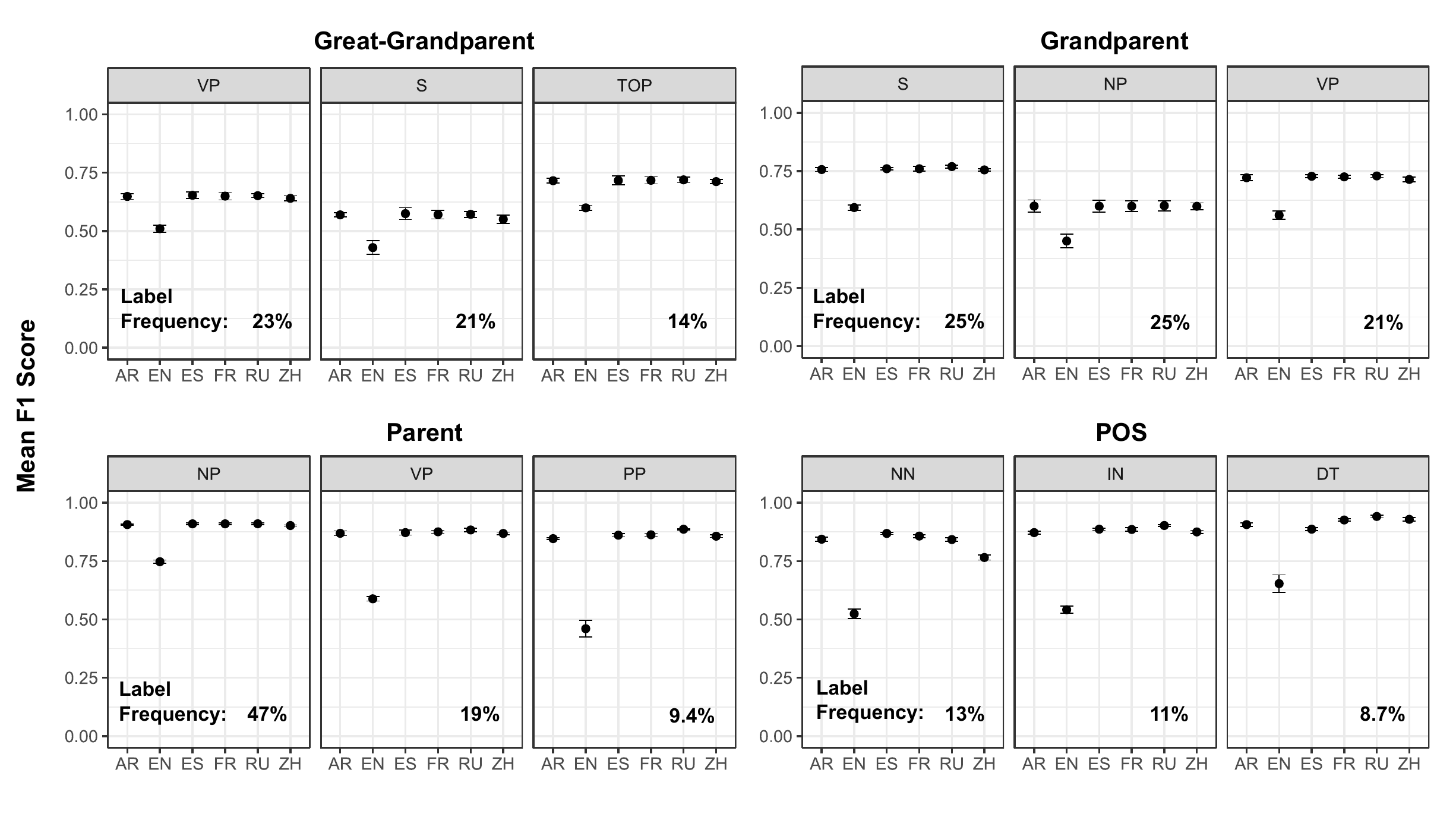}
    \caption{Mean F1 scores (based on 20 feedforward models) for individual constituent label predictions, treating each prediction task as a binary classification task.
    Bars indicate two standard deviations from the mean.
    We display the three most frequent labels for each task, comparing across all target languages.
    Each label’s frequency in the CoNLL-2012 test set is displayed on its corresponding plot.}
    \label{fig:F1Scores}
\end{figure*}

Individual constituent label F1 scores correlated extremely highly between non-English target languages (all pairwise Pearson correlations $r > 0.93$ for the POS task; $r > 0.98$ for the parent task; $r > 0.99$ for the grandparent and great-grandparent tasks).
In other words, the models performed well or poorly on the same individual labels regardless of target language.
Figure \ref{fig:F1Scores} shows individual constituent label F1 scores for each NMT target language, displaying the three most frequent labels for each constituent label task.
Similar to the overall accuracy scores, raw differences in F1 scores were small between non-English target languages.

In particular, the similar F1 scores were not simply proportional to label frequencies.
For instance, all target languages performed similarly well when identifying noun grandparent constituents (25\% of grandparent
labels, F1 scores 0.59-0.60) and question-sentence grandparent
constituents (0.6\% of grandparent labels, F1 scores
0.55-0.61), despite over a 20\% difference in corresponding
label frequencies.\footnote{There was a loose correlation between F1 scores and label frequencies, but this correlation could not fully account for the similarity of F1 scores across target languages.}
Similar F1 scores across non-English target languages suggest that NMT encoders encode very similar syntactic information regardless of target language.

\begin{table}
    \centering
    \begin{tabular}{|c|c|c|c|c|c|}
        \hline
        \multicolumn{6}{|c|}{Tokenized BLEU} \\
        \hline
        AR & EN & ES & FR & RU & ZH \\
        \hline
        37.3 & 99.9 & 56.3 & 44.8 & 37.8 & 24.9 \\
        \hline
        \multicolumn{6}{c}{ } \\
        \hline
        \multicolumn{6}{|c|}{Detokenized BLEU} \\
        \hline
        AR & EN & ES & FR & RU & ZH \\
        \hline
        38.0 & 100.0 & 56.3 & 44.5 & 37.4 & \\
         \hline
    \end{tabular}
    \caption{BLEU scores before and after detokenizing the NMT translations for the UN test set.
    The detokenized BLEU score was not computed for Chinese because words were generally not separated by spaces in the Chinese dataset.}
    \label{tab:Bleu}
\end{table}

\paragraph{Translation quality still varied.}
Despite similar syntactic information encoded across target languages, the NMT models exhibited a wide range of BLEU scores, as shown in Table \ref{tab:Bleu}.
This indicates that morphological and non-syntactic features have large impacts on translation performance.
For instance, inflectional morphology (e.g. verb conjugation and noun pluralization) has been found to account for differences in performance between languages in language modeling tasks \citep{cotterell-etal-2018-languages}, although these results vary depending on the metric used for morphological complexity \citep{mielke-etal-2019-kind}.
Because differences in translation performance could not be easily explained using encoded syntactic information alone, it seems likely that the NMT models were either unable to extract more syntactic information from the training data or that the models did not find additional syntactic information to be useful.

\subsection{Linguistic Analysis of Errors}
\label{sec:LingAnalysis}
To gain a better understanding of how NMT encoders extract syntax, we conducted a qualitative analysis of sentences for which the constituent label prediction models exhibited high error rates.

\paragraph{Selection of sentences}
We selected sentences based on the great-grandparent constituent label task because this task exhibited the highest accuracy scores above the baseline, indicating a large amount of learned syntax.
There were high pairwise correlation scores for per-sentence great-grandparent constituent label accuracies between all non-English target languages (all Pearson correlations $r > 0.85$), so we selected sentences simply based on their average constituent label accuracy across the five non-English target languages.

We considered the 50 complete sentences with the highest average great-grandparent constituent accuracies and the 50 complete sentences with the lowest average great-grandparent constituent accuracies.\footnote{
Sentences were marked as ``complete'' by a native English speaker. We considered only sentences from text sources (e.g. not transcribed conversational speech).
}
The top 50 sentences all had average great-grandparent accuracies above 90\%, and the bottom 50 sentences all had accuracies below 35\%.
Linguistic patterns found in the top and bottom 50 sentences are compiled in Table \ref{tab:LingFeatures}.

\begin{table}
    \centering
    \begin{tabular}{|p{2.95cm}|p{1.72cm}|p{1.72cm}|}
        \hline
        \textbf{Feature} & \textbf{Top 50} & \textbf{Bottom 50} \\
        & \textbf{sentences} & \textbf{sentences} \\
        \hline
         Average length & 9.3 words & 21.7 words \\
         \hline
         \raggedright Average great-grandparent constituent label accuracy & 0.949 & 0.310  \\
         \hline
         \raggedright Question sentences & 2 & 10 \\
         \hline
         Infinitive phrases & 26 & 5 \\
         \hline
         \raggedright Sentences with negation & 13 & 4 \\
         \hline
         \raggedright Sentences containing a null copula or appositive & 0 & 16 \\
         \hline
         \raggedright Embedded sentences (excluding infinitives) & & \\
         \textcolor{white}{\rule{0.3cm}{0.45cm}} \raggedright $\circ$ Head before & 9 & 5 \\
         \rule{0.3cm}{0cm} \raggedright $\circ$ Head after & 0 & 10 \\
         \hline
    \end{tabular}
    \caption{Linguistic features in the top and bottom 50 sentences, selected based on great-grandparent constituent label accuracies per sentence.}
    \label{tab:LingFeatures}
\end{table}

\paragraph{NMT encoders relied on explicit cues.}
The bottom 50 sentences contained a disproportionate number of null features.
These features omit words or morphemes that would indicate syntactic structure in a sentence.
For instance, null copulas omit forms of the verb ``to be,'' as in the sentence ``He pronounced the homework [was] finished.''
Appositives, where two noun phrases are placed one after another to describe the same entity (e.g. ``Grant, the star baker''), serve as relative clauses with the usual explicit syntactic cues omitted (e.g. ``Grant, [who is] the star baker'').
Of the bottom 50 sentences, 16 contained at least one null copula or appositive; the top 50 sentences contained none of either feature.
This suggests that when generating encoder representations, NMT models typically do not identify syntactic structures based on non-explicit cues.

However, the models performed well on complex syntactic structures containing explicit morphosyntactic cues.
They performed well on sentences containing infinitives (e.g. ``to eat'' or ``to pillage'') and negation (e.g. ``I did not eat''), exhibiting far more of these features in the top 50 sentences than in the bottom 50 sentences (see Table \ref{tab:LingFeatures}).
Both infinitives and negation have clear morphosyntactic cues indicating sentence structure.  The ``to'' in each infinitive clearly introduces the infinitized verb, and the word ``not'' before a verb clearly indicates a negated clause.  These results suggest that NMT encoders rely on explicit morphosyntactic cues to extract syntactic structure from sentences.

\paragraph{NMT encoders recognized embedded sentences.}
In fact, the NMT encoders were able to use morphosyntactic cues to identify embedded sentences.
An embedded sentence appears within another phrase (e.g. within the verb phrase ``said that [sentence]'').
The phrase head which introduces an embedded sentence can appear before or after the embedded sentence (e.g. ``Alex said [sentence]'' versus ``[sentence], said Alex'').
Because the NMT encoders were forward-directional RNNs, they could not be expected to recognize embedded sentences where the corresponding phrase head appeared after the embedded sentence.
However, the models performed well on many sentences where the phrase head appeared before the embedded sentence, exhibiting nine such structures in the top 50 sentences (see Table \ref{tab:LingFeatures}).
In many of these sentences, the head and complementizer (e.g. ``said that'' or ``dogs that'') clearly indicate the beginning of an embedded sentence.

Interestingly, the NMT encoders were often able to recognize embedded sentences even when there was a null complementizer introducing the embedded sentence, such as ``that'' omitted in ``The dog wished [that] he was taller.''
Of the nine embedded sentences in the top 50 sentences, six had a null complementizer.
This result may partially be explained by verb bias, the tendency for certain verbs to be followed by particular types of phrases \citep{garnsey-etal-1997-contributions}.
For instance, the verb ``prove'' is more often followed by a sentence complement (e.g. ``proved [that] the criminal was lying'') than a direct object (e.g. ``proved the theorem'').
People are more likely to omit complementizers when the head verb biases heavily towards a sentence complement \citep{ferreira-schotter-2013-verb}; in these cases, the verb itself serves as a syntactic cue for the upcoming embedded sentence.
Of the six null complementizers in the top 50 sentences, five followed a sentence-complement-biased verb.
Then, it appears that NMT encoders are able to recognize embedded sentences using a combination of verb bias and explicit complementizers.

\section{NMT Syntax vs. Other Models}
The similarity of syntactic information in NMT encoder representations across target languages could suggest that regardless of target language, a similar amount of syntactic information is helpful for translation.
However, it is also possible that the structure of the constituent label task limited the syntactic information the encoders could represent, as predicting a label based only on a partial sentence is an inherently ambiguous task.
A third alternative is that the RNN encoder architectures limited the information preserved in each representation.

To further explore how well the NMT encoders extracted syntactic information from raw sentences, we compared their constituent label prediction performance to two alternative models: an RNN trained directly for the constituent label task, and a probabilistic context-free grammar (PCFG) parser.
In contrast to the NMT encoders, the RNN can learn representations that are best suited for retaining syntax; like the NMT encoders, it sees one word at a time.
The PCFG is trained with complete syntactic information for partial sentences, and its prediction task is an entire hierarchical structure, rather than a single type of label.
These comparisons can show whether there are syntactic features that are predictable but systematically missed by the NMT encoder representations.

\subsection{Directly-Trained RNNs}
\paragraph{RNN models}
We trained unidirectional four-layer LSTM models with 500 dimensions to directly predict constituent labels (POS, parent, grandparent, great-grandparent) when provided a sentence stopping at a given word.
These RNNs were trained on the CoNLL-2012 development dataset (the same dataset as the feedforward models based on NMT encoder representations in Section \ref{sec:PredictionModels}).
To account for variance in RNN training, we trained 10 RNNs for each constituent label task, and each RNN was trained until it completed 10 consecutive epochs without improvement on the evaluation dataset.

\begin{figure}
    \centering
    \includegraphics[width=7.7cm]{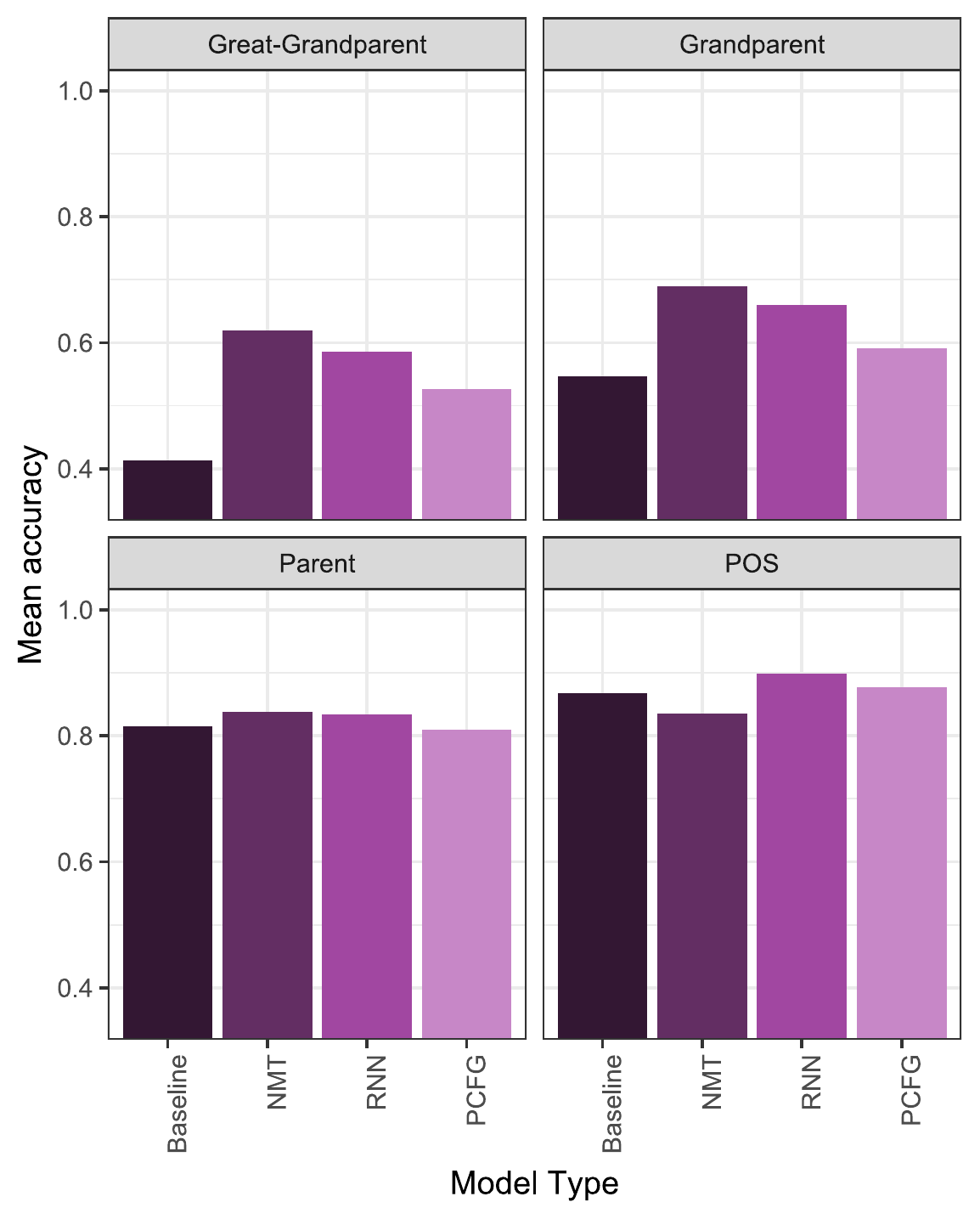}
    \caption{Average accuracies on the constituent label prediction tasks for all four types of model.}
    \label{fig:ModelAccuracies}
\end{figure}

\paragraph{NMT representations outperformed the RNNs.}
Average accuracies for the RNN models in each constituent label task are shown in Figure \ref{fig:ModelAccuracies}, compared with the feedforward models trained from NMT encoder representations.
Surprisingly, the RNN models trained directly for the constituent label tasks performed worse than the NMT encoder representation models for the parent, grandparent, and great-grandparent constituent tasks.
The NMT encoder representations' improvement over the other models increased consistently as the constituent labels moved higher in the syntax tree (i.e. the NMT encoders exhibited the greatest advantage in the great-grandparent constituent task).

Because the RNNs had the same architecture as the NMT encoders, it is likely that the directly-trained RNNs were limited by the amount of training data provided (about 160,000 examples).
The NMT encoder representations would be able to rely more heavily on patterns learned during NMT training and thus would be able to make better use of the limited training data for the constituent label prediction tasks.
It is also possible that the hyperparameters used for the NMT encoders were not optimal for the directly-trained RNNs.
That said, the NMT encoder representations' high performance on the constituent label tasks supports existing literature finding that NMT encoder representations contain information useful for a variety of natural language tasks (\citealp{Eriguchi2018ZeroShotCC}; \citealp{McCann2017LearnedIT}).

\paragraph{The RNNs and NMT encoded similar syntax.}
Next, to assess whether the directly-trained RNNs learned different syntactic information from the NMT encoders, we compared the RNN and the NMT encoder representations' performance on individual sentences.
We primarily considered great-grandparent constituent accuracies, the task for which all models performed most above the baseline.

\begin{figure}
    \centering
    \includegraphics[width=7.7cm]{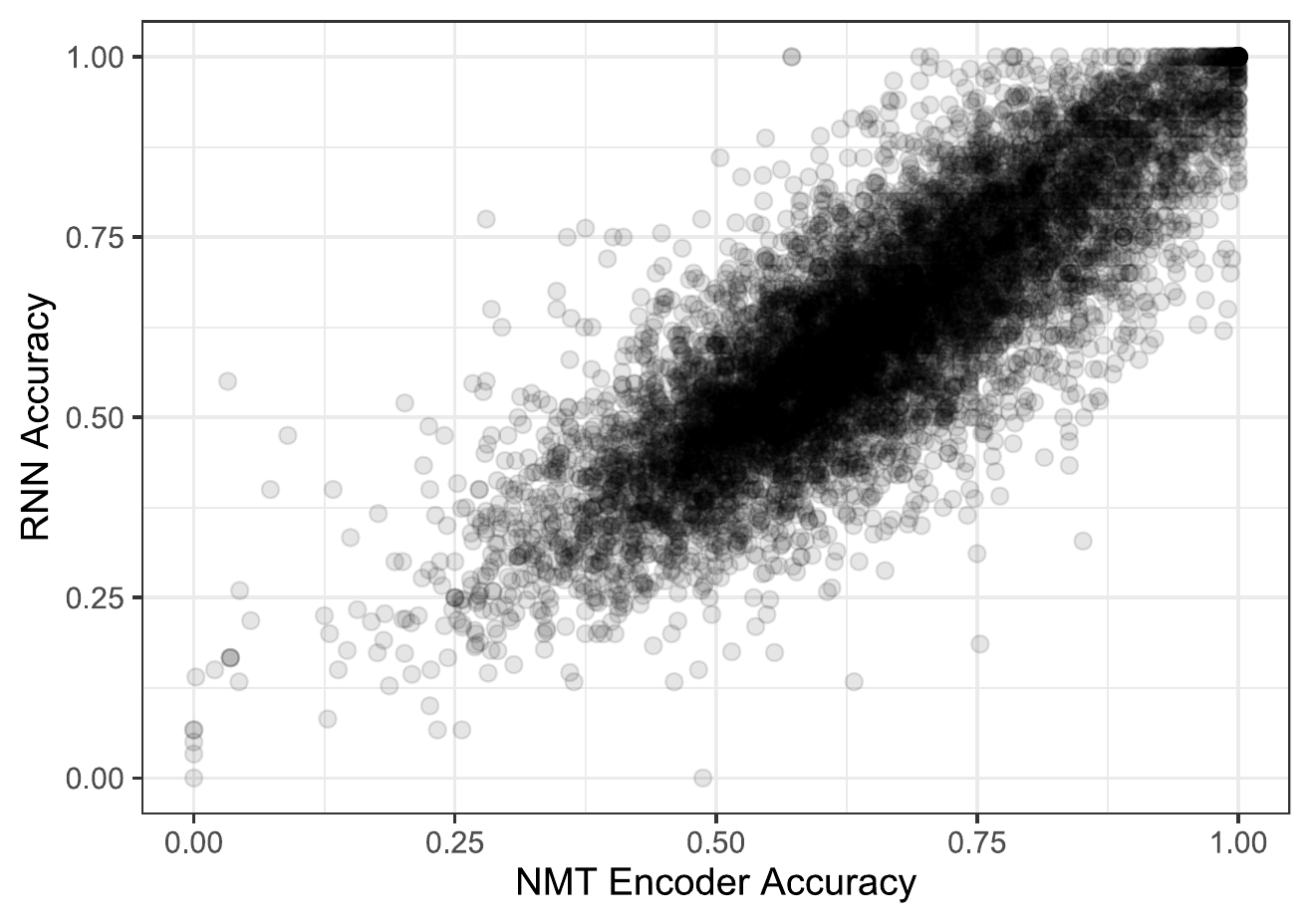}
    \caption{Mean great-grandparent constituent label accuracies per sentence for the NMT encoder-based models and the directly-trained RNNs.
    Each dot represents a sentence.}
    \label{fig:Correlation1}
\end{figure}

For each sentence of length at least three, we considered the mean great-grandparent constituent label accuracy, averaging across all non-English target languages for the NMT encoder accuracies.
Figure \ref{fig:Correlation1} shows the correlation between per-sentence accuracies from the NMT encoder representation models and the directly-trained RNN models.
There was a high degree of correlation between the two types of models (Pearson correlation $r=0.84$), indicating that the directly-trained RNNs learned similar syntactic information to the NMT encoders.

\begin{figure}
    \centering
    \includegraphics[width=7.7cm]{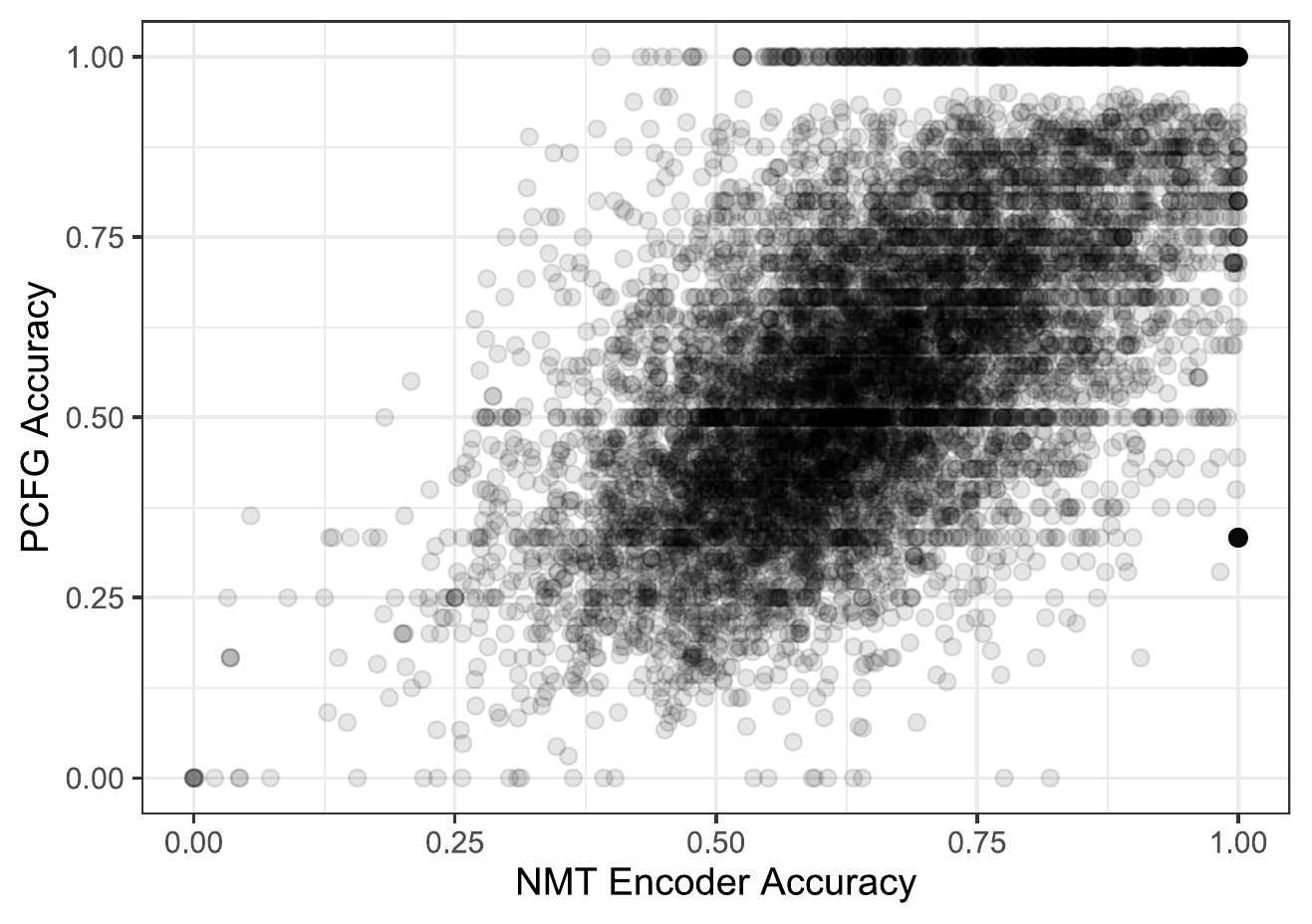}
    \caption{Mean great-grandparent constituent label accuracies per sentence for the NMT encoder-based models and the PCFG parser.}
    \label{fig:Correlation2}
\end{figure}

\subsection{PCFG Parser}
It may be that the directly-trained RNNs and the NMT encoders learned similar syntactic information because they both used the same RNN architecture.
Therefore, we tested constituent label performance when using the probabilistic context-free grammar (PCFG) syntactic parser provided by Stanford NLP \citep{klein-manning-2003-accurate}.
We trained the PCFG on parse trees of partial sentences stopping at each word in the CoNLL-2012 development dataset, the same dataset used to train the RNN-based models.
While the PCFG was not trained specifically for the constituent label prediction task, its explicit syntactic architecture (encoding a context-free grammar) provides a useful contrast to the RNN-based models.

\paragraph{The PCFG encoded different syntax.}
The PCFG's constituent label accuracies are shown in Figure \ref{fig:ModelAccuracies}, along with the RNN and NMT encoder representation accuracies.
As expected, because the PCFG was not trained specifically for the constituent label prediction task, the PCFG had slightly lower accuracies than the RNN-based models.
However, the PCFG exhibited interesting patterns when considering its performance on individual sentences.

\begin{table}
    \centering
    \begin{tabular}{|c|c|c|c|}
        \cline{2-4}
         \multicolumn{1}{c|}{} & RNN & Baseline & PCFG \\
         \hline
         NMT & 0.84 & 0.60 & 0.61 \\
         \hline
         RNN & & 0.62 & 0.59 \\
         \hline
         Baseline & & & 0.42 \\
         \hline
    \end{tabular}
    \caption{Pairwise Pearson correlations for per-sentence great-grandparent constituent label accuracies, computed between all four types of model.}
    \label{tab:Correlations}
\end{table}

As with the other models, the PCFG's mean great-grandparent constituent label accuracies were considered for each sentence of length at least three.
Figure \ref{fig:Correlation2} (comparing the PCFG with the NMT encoder representations) can then be compared to Figure \ref{fig:Correlation1} (comparing the directly-trained RNNs with the NMT encoder representations).
The two plots indicate that the PCFG performed substantially differently from the RNN-based models.
Notably, there is a set of sentences for which the PCFG obtained perfect accuracy while the NMT encoders had substantially lower accuracies (demonstrated by the horizontal line of dots at the top of Figure \ref{fig:Correlation2}).
Both RNN-based models' accuracies correlated approximately the same amount with the baseline (most-frequent tag per word) model as with the PCFG; all correlations between models are shown in Table \ref{tab:Correlations}.

Furthermore, for the worst 50 sentences for the NMT encoder representations (the sentences found in Section \ref{sec:LingAnalysis}), the PCFG performed 9\% better than the NMT encoder representation models and 6\% better than the directly-trained RNN models, despite an overall 7-9\% lower accuracy than both RNN-based models.
This suggests that PCFGs can perform well on specific sentences that RNNs perform poorly on; for instance, PCFGs may be less reliant on explicit morphosyntactic cues.
The PCFG's high performance on these specific sentences explains results finding that explicit syntactic information provides improvements to NMT systems even though NMT systems already implicitly encode syntax (\citealp{chen-etal-2017-improved}; \citealp{chiang-etal-2009-11001}; \citealp{li-etal-2017-modeling}; \citealp{wu-etal-207-improved}).

\section{Discussion}
\paragraph{NMT syntax is independent of target language.}
We found that NMT encoders learn similar source syntactic information regardless of target language, consistently outperforming RNNs trained specifically for the constituent label prediction task.
These results help explain the success of NMT encoder representations in cross-task transfer learning, and they open up further questions regarding the extent of similarity between NMT encoder representations across target languages.

For instance, \citet{schwenk-douze-2017-learning} found that multilingual NMT encoder representations cluster more based on semantic than syntactic similarity, indicating that semantic information may play a more prominent role than syntax in machine translation.
Across target languages, \citet{poliak-etal-2018-evaluation} found inconsistencies for which target language's representations resulted in the best performance on semantic understanding tasks.
This could suggest that semantic information in NMT encoder representations is also similar across target languages.

\paragraph{RNNs learn limited syntax.}
Both the NMT encoders and the directly-trained RNNs relied on explicit morphosyntactic cues to extract syntactic information from sentences.
This result aligns with findings that RNNs rely on syntax heuristics to obtain high performance on tasks \citep{mccoy-etal-2019-right}, performing poorly on sentences requiring knowledge of complex syntactic structures (\citealp{linzen-etal-2016-assessing}; \citealp{marvin-linzen-2018-targeted}).
NMT encoders specifically have been found not to encode fine-grained syntactic information \citep{shi-etal-2016-string}.
These limitations can be partially overcome by training an RNN model for a variety of different tasks \citep{enguehard-etal-2017-exploring}; alternatively, we found that a PCFG syntactic parser encoded significantly different syntactic information from RNN-based models, performing well on many sentences for which RNNs performed poorly.

In some ways, the RNNs' reliance on explicit syntactic cues is similar to sentence processing in people.
Many sentences are syntactically ambiguous before they are completed (notably garden-path sentences such as ``The horse raced past the barn fell''), and people generally re-evaluate upon reading the disambiguating feature (\citealp{frazier-rayner-1982-correcting}; \citealp{qian-etal-2018-comparison}).
Thus, it may be implausible for an online system to identify non-explicit syntactic features given only partial sentences.
Compounding this problem, RNNs are unable to re-evaluate past inputs and hidden states upon reading disambiguating words.
The successes of bidirectional and Transformer models (\citealp{devlin-etal-2019-bert}; \citealp{peters-etal-2018-deep}; \citealp{vaswani-etal-2017-attention}) may be due partially to their ability to combine later information with representations of earlier words.
Indeed, contextual word representations generated by these bidirectional models have been found to encode significant syntactic information \citep{peters-etal-2018-dissecting}; future work could study whether bidirectional architectures are less reliant on explicit morphosyntactic cues.

\section{Conclusion}
In this work, we found that NMT encoder representations across target languages encode similar source syntax, and this syntax is comparable to the syntax learned by RNNs trained directly on syntactic tasks.
However, explicit syntactic architectures may be necessary for tasks requiring fine-tuned syntactic parses.
Our results have many implications in transfer learning and multilingual sentence representations: a better understanding of the information contained in sentence representations provides necessary insight into the tasks these representations can be used for.

\section*{Acknowledgments}
We would like to thank Cherlon Ussery for helpful linguistic perspectives on our results, and the Carleton College Cognitive Science Department for making this work possible.

\bibliography{anthology,acl2020,other}
\bibliographystyle{acl_natbib}

\appendix

\end{document}